\documentclass[10pt,twocolumn,letterpaper]{article}
\usepackage{geometry}
\usepackage{cvpr}              

\usepackage[dvipsnames]{xcolor}
\usepackage{multirow}
\usepackage{graphicx}
\usepackage{tabularray}
\usepackage{rotating}
\usepackage{makecell}
\usepackage{array}

\definecolor{cvprblue}{rgb}{0.21,0.49,0.74}
\usepackage[pagebackref,breaklinks,colorlinks,citecolor=cvprblue]{hyperref}

\def\paperID{33} 
\def\confName{CVPR}
\def\confYear{2024}

\title{Understanding the Impact of Training Set Size on Animal Re-identification}

\author{Aleksandr Algasov, Ekaterina Nepovinnykh, Tuomas Eerola, Heikki K\"{a}lvi\"{a}inen \\
		Computer Vision and Pattern Recognition Laboratory (CVPRL)\\
	 LUT University, Lappeenranta, Finland\\
		{\tt\small firstname.lastname@lut.fi}
   \and
    Charles V. Stewart  \\
    Department of Computer Science \\
    Rensselaer Polytechnic Institute, Troy, NY, USA \\
    {\tt\small stewart@rpi.edu}
    \and
    Lasha Otarashvili, Jason A. Holmberg  \\
    Wild Me, Conservation X Labs\\
    Portland, OR, USA \\
    {\tt\small firstname@conservationxlabs.org}
}

\begin{document}
\maketitle

\begin{abstract}

Recent advancements in the automatic re-identification of animal individuals from images have opened up new possibilities for studying wildlife through camera traps and citizen science projects. Existing methods leverage distinct and permanent visual body markings, such as fur patterns or scars, and typically employ one of two strategies: local features or end-to-end learning. In this study, we delve into the impact of training set size by conducting comprehensive experiments across six different methods and five animal species. While it is well known that end-to-end learning-based methods surpass local feature-based methods given a sufficient amount of good-quality training data, the challenge of gathering such datasets for wildlife animals means that local feature-based methods remain a more practical approach for many species. We demonstrate the benefits of both local feature and end-to-end learning-based approaches and show that species-specific characteristics, particularly intra-individual variance, have a notable effect on training data requirements.

\end{abstract}

\section{Introduction}
\label{sec:intro}

Camera trapping and crowdsourcing have become prevalent tools for collecting extensive image data to study wildlife populations. This has created a demand for computer vision methods to automate the analysis of the image data. One of the most important tasks to be automated is the re-identification of individual animals, as this provides the basis for animal behavior analysis and more accurate population size estimates. 

A number of methods have been proposed for animal re-identification~\cite{hotspotter,zhelezniakov2015segmentation,chehrsimin2018automatic,nepovinnykh2018identification,nepovinnykh2020siamese,nepovinnykh2024norppa}. Generally, the methods operate by quantifying various visual markers that are both permanent and unique to each individual. These include fur, skin, and feather patterns, ear and fin shapes, as well as facial features.
The existing methods can be divided into three categories: (1) handcrafted species-specific approaches, (2) local feature-based approaches, and (3) end-to-end deep learning approaches. The first category consists of methods that quantify handpicked features such as the ear or fin shape (see e.g.~\cite{hughes2017automated}) of a specific animal species. Such methods cannot be generalized to a wide variety of different species; therefore, they are out of the scope of this study.

The methods in the second category use local features to detect and to describe interesting regions in the body markings. They can be further divided into methods utilizing traditional handcrafted local features such as SIFT~\cite{lowe2004distinctive}, and methods employing learnable local features, such as CNN-based HardNet~\cite{HardNet2017} and SuperPoint~\cite{detone2018superpoint}. The local feature-based methods also differ in how the local feature descriptors are utilized to quantify the body marking similarity for re-identification. The most commonly used approaches are (1) geometry-based matching methods using descriptor matching and RANSAC-like methods to analyze the geometric consistency of the matching regions~\cite{hotspotter,immonen2023combining}, and (2) feature aggregation-based methods that use, for example, the Fisher vector to aggregate features into a single embedding vector allowing efficient database search for the most similar pattern~\cite{chelak2021eden, nepovinnykh2022matching, nepovinnykh2023re, nepovinnykh2024norppa}. CNN-based local features combined with feature aggregation have shown to produce representative pattern feature embeddings and high re-identification accuracy~\cite{nepovinnykh2024norppa,nepovinnykh2024alfreid}. At the same time, pre-trained or handcrafted local features make it possible to apply these methods even when there is no training data available.

In the third category, the pattern embeddings are learned in an end-to-end manner using deep metric learning techniques. Both CNN~\cite{schneider2022similarity,moskvyak2021keypoint} and transformer-based~\cite{vcermak2024wildlifedatasets,jiao2024toward} models have been proposed for animal re-identification. These methods make it possible to learn necessary features for re-identification from the images of the full animal without the need to locate the distinctive regions from the body marking. The end-to-end learning-based methods tend to surpass local feature-based methods given a sufficient amount of good-quality training data. However, gathering large datasets of wildlife animals remains a challenge.

This study investigates the impact of training set size on animal re-identification across different methods. We seek to determine how the number of training images influences method selection and the extent to which animal species affect training data requirements. Our focus is the “small number of images” scenario because, in real life, the available annotated datasets are very limited in terms of the number of images, individuals, and the variety of images per individual. In this study, we conduct extensive experiments on five animal species using six re-identification methods including both local feature and end-to-end learning-based methods.
 

\section{Methods} 
\label{sec:methods}

For this study, we selected six distinct methods. The local feature-based methods included are Hotspotter~\cite{hotspotter} and Aggregated Local Features for Re-Identification (ALFRE-ID)~\cite{nepovinnykh2024alfreid}. The former utilizes handcrafted local features, while the latter employs pre-trained CNN-based local features and feature aggregation. The end-to-end learning-based methods included are the CNN-based MiewID~\cite{miewid}, as well as the transformer-based TransReID~\cite{transreid}, MegaDescriptor~\cite{vcermak2024wildlifedatasets}, and BioCLIP~\cite{stevens2024bioclip}. TransReID is a general-purpose re-identification method originally developed for humans and vehicles, whereas MegaDescriptor was specifically developed for animal re-identification, and BioCLIP is a general biology vision model.

\subsection{Hotspotter}
\label{ssec:hotspotter}

Hotspotter~\cite{hotspotter} is a SIFT-based re-identification algorithm that uses viewpoint-invariant descriptors 
and a scoring mechanism that emphasizes the most distinctive keypoints or “hot spots” on a body marking. Moreover, a RANSAC-based spatial reranking is applied to ensure the geometric consistency of the point correspondences.

\subsection{ALFRE-ID}
\label{ssec:alfreid}

ALFRE-ID is an animal re-identification framework that 
starts with tonemapping and animal segmentation followed by pelage pattern extraction to emphasize the body marking essential for re-identification and to eliminate non-relevant elements such as background and illumination. 
Pre-trained CNN-based local feature extractors, such as HardNet~\cite{HardNet2017} or DISK~\cite{tyszkiewicz2020disk}, are used to extract local features that are then aggregated using Fisher vectors~\cite{perronnin2007fisher, perronnin2010large} to create a representative embedding of an individual's pattern.
Candidate matches are obtained by searching the most similar Fisher vectors from the database, and finally spatial reranking is performed similarly to Hotspotter. The main advantage of this method is that it does not need to be trained on the target dataset to perform well; instead it relies on pre-trained feature extractors. 
It should be noted that the ALFRE-ID framework does not specify the CNN-based local features, but they can be selected based on the animal species. These include HessAffnet~\cite{AffNet2018} for detection and HardNet~\cite{HardNet2017} for description, or DISK~\cite{tyszkiewicz2020disk} for both detection and description.

\subsection{MiewID}
\label{ssec:miewid}

MiewID~\cite{miewid} is a part of a WildBook project~\cite{berger2017wildbook}. It is an end-to-end learning-based re-identification method that utilizes the ArcFace loss~\cite{deng2019arcface} with an EfficientNet CNN architecture~\cite{tan2019efficientnet} as a backbone.

\subsection{TransReID}
\label{ssec:transreid}

TransReID is a transformer-based re-identification method originally developed for persons and vehicles~\cite{transreid}. It utilizes a sequence of image patches (tokens) and transformer layers for feature extraction~\cite{bagoftricks}. Moreover, a Jigsaw Patch Module that shifts and shuffles patch embeddings is applied to generate more robust features, and this way, to address occlusions and misalignments. The model is trained using the ID loss (the cross-entropy loss without label smoothing) and the triplet loss.

\subsection{MegaDescriptor}
\label{ssec:megadescriptor}

In~\cite{vcermak2024wildlifedatasets}, a toolkit for animal re-identification is introduced. Termed Wildlife Tools, this methodology encompasses a range of backbones, along with training and matching strategies. In this study, MegaDescriptor refers to the most effective model from the original publication, namely the Swin Transformer~\cite{liu2021swin} trained with the Arcface loss. In contrast to TransReID, MegaDescriptor was specifically developed for animal re-identification. 

\subsection{BioCLIP}
\label{ssec:bioclip}

BioCLIP is a vision model for general organismal biology~\cite{stevens2024bioclip}. It is based on OpenAI's ViT-B/16 CLIP model and trained on TreeOfLife-10M~\cite{treeoflife_10m}. The vision transformer encoder has lower input image resolution than MegaDescriptor, but due to nature of the TreeOfLife-10M dataset, it is expected to outperform general domain transformers with similar architecture in biology-specific tasks. For re-identification purposes, the internal vision transformer of the BioCLIP model was used to generate embeddings of images, which were then used to find the closest individual in the embeddings space.

\section{Experiments}
\label{sec:exp}

We studied the performance of the methods on multiple datasets and their scaled-down variants. 

\subsection{Data}
\label{ssec:datasets}

The experiments were conducted on five datasets, each representing a different species: Saimaa ringed seals (\emph{Pusa hispida saimensis}), whale sharks (\emph{Rhincodon typus}), Grevy's zebras (\emph{Equus grevyi}), Masai giraffes (\emph{Giraffa tippelskirchi}), and Beluga whales (\emph{Delphinapterus leucas}).

The Saimaa ringed seals dataset is a modified version of the SealID dataset~\cite{nepovinnykh2022sealid} where similar images were removed, and new images were added to enhance diversity. Saimaa ringed seals can be re-identified based on their unique and permanent pelage ringed patterns.

The Whale shark and Grevy's zebras datasets were provided by Wild Me~\cite{berger2017wildbook}. Masai giraffe dataset was provided by the Zoological Society of San Diego d/b/a San Diego Zoo Wildlife Alliance. These animals have unique biological traits that can be used for their re-identification: whale sharks have distinctive spot patterns on their skin, particularly around their dorsal fins and gills, Grevy's zebras have unique stripe patterns all over their bodies, and giraffes have spot patterns and neck markings.

\begin{figure}[!t]
    \centering
    \includegraphics[width=0.95\linewidth]{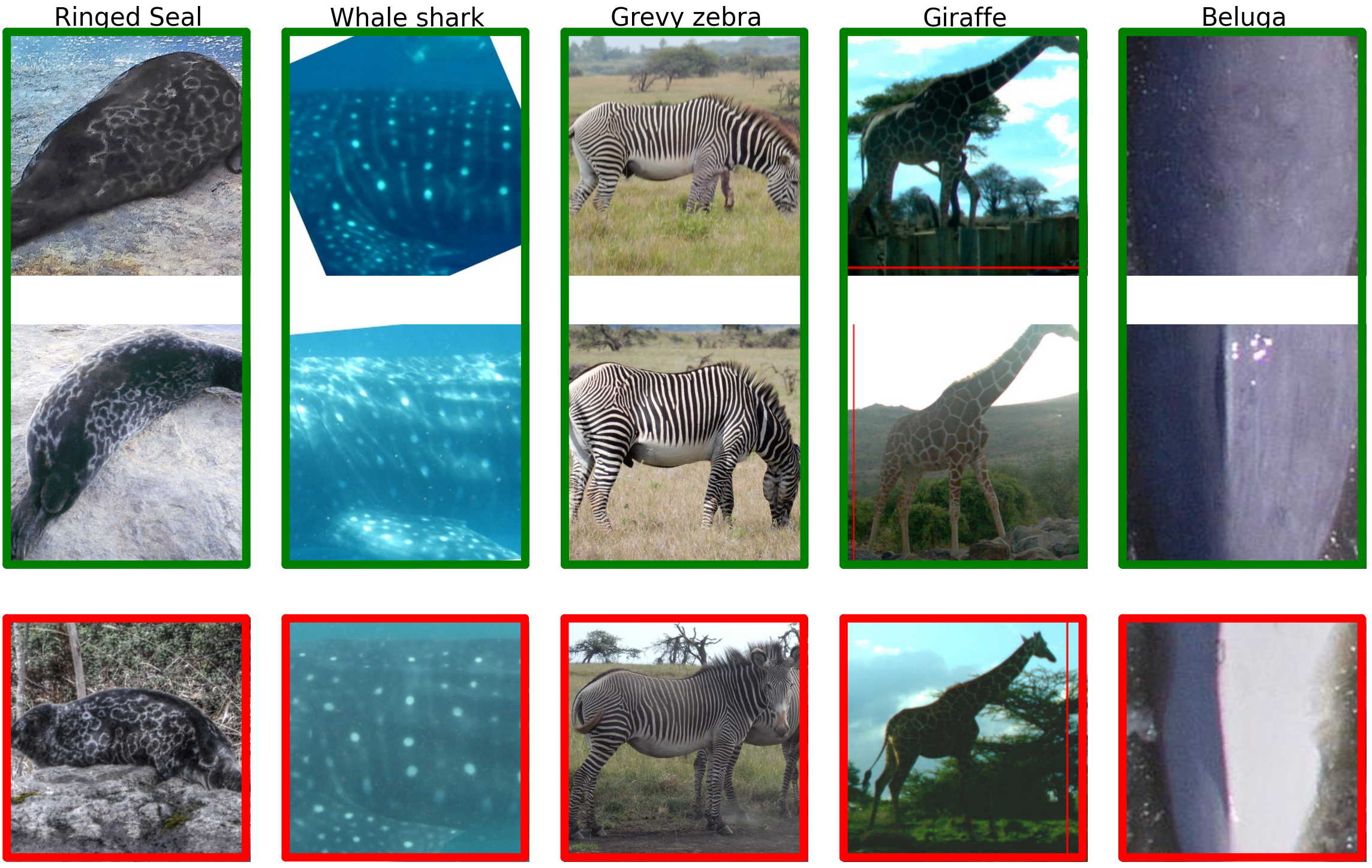}
    \caption{Sample images from datasets. Images connected with green borders are from the same individual, while images with red borders are from different individuals.}
    \label{fig:dataset_vis}
\end{figure}

The Beluga whales dataset is a publicly available dataset provided by LILA and Wild Me~\cite{beluga}. Beluga whales do not have a distinct skin pattern, but they can be re-identified using scars or pigmentation variations. Examples of images are presented 
in Fig.~\ref{fig:dataset_vis}. All dataset statistics are provided in Table~\ref{tab:data}.

\begin{table*}
\caption{Datasets and data splits used in the experiments.\label{tab:data}}
\vspace{-10pt}
\begin{footnotesize}
\centering
\begin{tblr}{
  colspec = {Q[h,wd=0.11\linewidth]Q[3]Q[3]Q[3]Q[3]Q[3]Q[3]Q[3]Q[3]Q[3]Q[3]Q[3]Q[3]Q[3]Q[3]Q[3]Q[3]},
  cell{1}{1} = {r=2}{},
  cell{1}{2} = {c=3}{},
  cell{1}{5} = {c=3}{},
  cell{1}{8} = {c=3}{},
  cell{1}{11} = {c=3}{},
  cell{1}{14} = {c=3}{},
  vlines,
  hline{1,4-9} = {-}{},
  hline{2-3} = {2-16}{},
}
                                  & Saimaa ringed seal &      &      & Whale shark &      &      & Grevy's zebra &      &     & Masai giraffe &      &     & Beluga whale &      &      \\
                                        & Train     & Test & Val  & Train      & Test & Val  & Train & Test & Val & Train   & Test & Val & Train  & Test & Val  \\
{Total~\\annotations}                       & 2000      & 480  & 462  & 4621       & 933  & 806  & 2535  & 690  & 631 & 2227    & 1471 & 583 & 2290   & 582  & 579  \\
{Number of \\ individuals   }                    & 277       & 146  & 146  & 306        & 119  & 120  & 215   & 345  & 174 & 455     & 695  & 89  & 286    & 188  & 187  \\
Average~annots per class & 14        & 4.8  & 4.6  & 15.1       & 7.8  & 6.7  & 11.8  & 2    & 3.6 & 5.0     & 2.1  & 6.6 & 8      & 3.1  & 3.1  \\
Unseen classes in test      & ~-        & 0.54 & 0.48 & -          & 0.55 & 0.57 & -     & 0.4  & 1.0 & -   & 1.0  & 1.0 &  -   & 0.5  & 0.45 

\end{tblr}
\end{footnotesize}
\end{table*}

\begin{table*}[t!]
\caption{Re-identification results. The highest top-1 accuracies for 0/500, 1000, 2000, and full size training sets are bolded.}
\begin{footnotesize}
\centering
\begin{tblr}{
  cell{1}{1,2} = {r=2}{valign=m},
  cell{1}{3,6,9,12,15} = {c=3}{},
  cell{3}{1-17} = {r=2}{valign=m},
  cell{5}{1-17} = {r=2}{valign=m},
  cell{7,11,15,19}{1} = {r=4}{},
  vlines,
  hline{1,3,5,7,11,15,19,23} = {-}{},
  hline{2} = {3-21}{},
  hline{8-10,12-14,16-18,20-22} = {2-21}{},
}
\begin{sideways}Method\end{sideways}         & \begin{sideways}Training\end{sideways}\begin{sideways}\hspace{6pt}data\end{sideways}\vspace{-3pt} & Saimaa ringed seal & & & Whale shark & & & Grevy's zebra & & & Masai giraffe & & & Beluga whale & & \\
& & Top-1 & Top-5 & mAP & Top-1 & Top-5 & mAP & Top-1 & Top-5 & mAP & Top-1 & Top-5 & mAP & Top-1 & Top-5 & mAP \\
\begin{sideways}Hotspotter~\cite{hotspotter}\end{sideways}\vspace{-3pt}
& 0 & 44.3 & 47.3 & 18.6 & 49.45 & 51.0 & 27.4 & \textbf{78.4} & 87.2  & 82.6 & \textbf{84.9} & 94.5 & 38.2 & 0 & 0 & 0 \\
& & & & & & & & & & & & & & & & \\
\begin{sideways}ALFRE-ID~\cite{nepovinnykh2024alfreid}\end{sideways}\vspace{-3pt}
& 0 & \textbf{53.9} & 56.25 & 33.2 & \textbf{73.5} & 77.4 & 49.1 & 69.6 & 77.8 & 40.5 & 72.3 & 79.4 & 26.1 & \textbf{40.4} & 50.6 & 22.2 \\
\begin{sideways}\end{sideways}
& & & & & & & & & & & & & & & & \\
\begin{sideways}MiewID~\cite{miewid}\end{sideways}         
& full & \textbf{43.8} & 55.4 & 30.8 & \textbf{95.0} & 97.3 & 90.3 & \textbf{99.1} & 99.4 & 99.2 & \textbf{84.3} & 91 & 87.5 & \textbf{69.8} & 79.4 & 67.5 \\
& 2000 & \textbf{43.8} &  55.4 & 30.8 & \textbf{96.5} & 98.2 & 92.0 & \textbf{98.7} & 99.6 & 99.1 & \textbf{76.4} & 87.0 & 81.2 & \textbf{46.7} & 65.1 & 41.8 \\
& 1000 & \textbf{40.0} & 52.3 & 27.8 & \textbf{86.9} & 93.8 & 72.2 & \textbf{95.8} & 97.8 & 96.7 & \textbf{28.4} & 53.8 & 40.5 & 30.8 & 46.7 & 25.5 \\
& 500 & 33.1 & 45.4 & 23.1 & 58.4 & 75.0 & 31.8 & 46.4 & 60.3 & 53.1 & 2.1 & 33.8 & 17.7 & 16.2 & 30.4 & 13.4 \\
\begin{sideways}TransReID~\cite{transreid}\end{sideways} 
& full & 34.4 & 44.0 & 22.3 & 93.2 & 96.7 & 84.3 & 45.2 & 60.4 & 52.8 & 18.0 & 38.7 & 28.2 & 38.5 & 59.8 & 34.4 \\
& 2000 & 37.7 & 48.3 & 25.5 & 52.2 & 76.3 & 32.3 & 42.2 & 56.4 & 49.4 & 15.0 & 34.7 & 25.1 & 42.6 & 62.9 & 36.7 \\
& 1000 & 34.8 & 46.9 & 25.3 & 43.9 & 69.1 & 23.9 & 27.1 & 40.7 & 34.0 & 17.3 & 32.2 & 25.1 & 25.9 & 46.7 & 23.7 \\
& 500 & 30.6 & 42.1 & 21.1 & 37.8 & 61.4 & 19.3 & 22.3 & 31.7 & 27.6 & 9.7 & 21.2 & 15.4 & 22.5 & 45.9 & 21.5 \\
\begin{sideways}MegaDescriptor~\cite{vcermak2024wildlifedatasets}\end{sideways}
& full & 32.9 & 43.3 & 22.9 & 86.7 & 93.7 & 74.7 & 40.6 & 53.9 & 47.3 & 15.7 & 27.9 & 22.4 & 44.2 & 61.5 & 39.7 \\
& 2000 & 32.9 & 45.4 & 23.4 & 45.8 & 72.0 & 28.0 & 32.0 & 44.3 & 38.5 & 14.5 & 26.7 & 21.3 & 43.0 & 63.9 & 39.4 \\
& 1000 & 30.4 & 42.5 & 21.9 & 32.5 & 57.6 & 16.7 & 23.6 & 32.3 & 28.8 & 11.0 & 20.4 & 16.4 & \textbf{33.5} & 52.1 & 29.0 \\
& 500 & 28.1 & 38.1 & 19.3 & 19.4 & 40.0 & 8.6 & 24.6 & 32.8 & 28.8 & 9.9 & 17.9 & 14.7 & 24.6 & 41.8 & 20.9 \\
\begin{sideways}BioCLIP~\cite{stevens2024bioclip}\end{sideways} 
& full & 29.4 & 41.5 & 19.5 & 86.3 & 95.2 & 75.3 & 32.5 & 48.6 & 40.9 & 18.9 & 36.0 & 28.2 & 23.0 & 43.3 & 22.8 \\
& 2000 & 29.4 & 41.5 & 19.5 & 65.5 & 84.4 & 45.2 & 29.7 & 44.1 & 36.9 & 19.2 & 37.4 & 28.6 & 25.6 & 45.5 & 22.1 \\
& 1000 & 27.1 & 38.5 & 17.9 & 40.9 & 65.4 & 24.1 & 23.0 & 33.5 & 28.7 & 18.6 & 35.9 & 27.9 & 18.0 & 38.7 & 16.2 \\
& 500 & 25.4 & 34.6 & 16.4 & 13.2 & 34.2 & 7.0 & 15.2 & 23.0 & 19.5 & 9.1 & 19.6 & 15.2 & 15.6 & 32.1 & 14.5
\end{tblr}
\end{footnotesize}
\label{tab:results}
\end{table*}

\subsection{Description of experiments}

We created several scaled-down versions of the training sets for each species, comprising 2000, 1000, and 500 images each, and evaluated the accuracy of the methods when trained on these sets. The scaled-down training sets were created by iteratively removing annotations from classes with the highest counts until the desired total was reached.Test and validation sets were created using an optimizing split objective function. The objective function minimizes the difference between the proportions of training and unseen data classes while maintaining the specified train ratio and unseen ratio. For all datasets, except Masai giraffe, the training ratio was set to 0.7, meaning the test and validation set ratios were 0.15 each. For Masai Giraffe, the original split was used. For Grevy's zebra the unseen ratio was set to 1.0, indicating that all individuals in the test subset were not seen during training. For Saimaa Ringed Seals, whale sharks, and beluga whales, the unseen ratio was set to 0.5, meaning that 50\% of all individuals in the test subset were presented in the training subset. No images presented in the test were seen during training. Note that Hotspotter and ALFRE-ID do not require training.

Hotspotter was applied with default parameters to test the selected datasets. 
The ALFRE-ID framework supports various feature extractors. For the seal dataset, HessAffNet~\cite{AffNet2018} and HardNet~\cite{HardNet2017} were used as feature detector and descriptor respectively. The images were preprocessed by removing the background and extracting the pelage pattern. For the other datasets, DISK~\cite{tyszkiewicz2020disk} features were used. For the zebra and giraffe datasets, the background was removed. The Beluga whale data did not require preprocessing as the images were pre-cropped and resized.
The codebooks were created from the test datasets.

For MiewID, standard training parameters were used except for the image size which was set to $480\times480$.

For TransReID, the ViT-B/16 backbone pre-trained on ImageNet-21K and ImageNet-1K was used. The model was fine-tuned on each dataset using the Adam optimizer for 100 epochs, and the cosine decay with a restarts scheduler, and $224\times224$ input size images.

For MegaDescriptor, the Swin-L-384 backbone pre-trained on WildlifeDatasets~\cite{vcermak2024wildlifedatasets} was chosen. To avoid the problem of images from the datasets under consideration being partially included into the train data of MegaDescriptor, the model was re-trained using open-sourced code provided by the WildlifeDatasets, with exclusion of the five datasets described in Section ~\ref{ssec:datasets}. The fine-tuning was done with the Arcface loss function, the Adam optimizer, and the cosine annealing learning rate schedule over 50 epochs.

For BioCLIP fine-tuning was done only on internal vision model, with otherwise the same training setting as for MegaDescriptor.

\section{Results and discussion}

\begin{figure}
    \centering
    \includegraphics[width=1.0\linewidth]{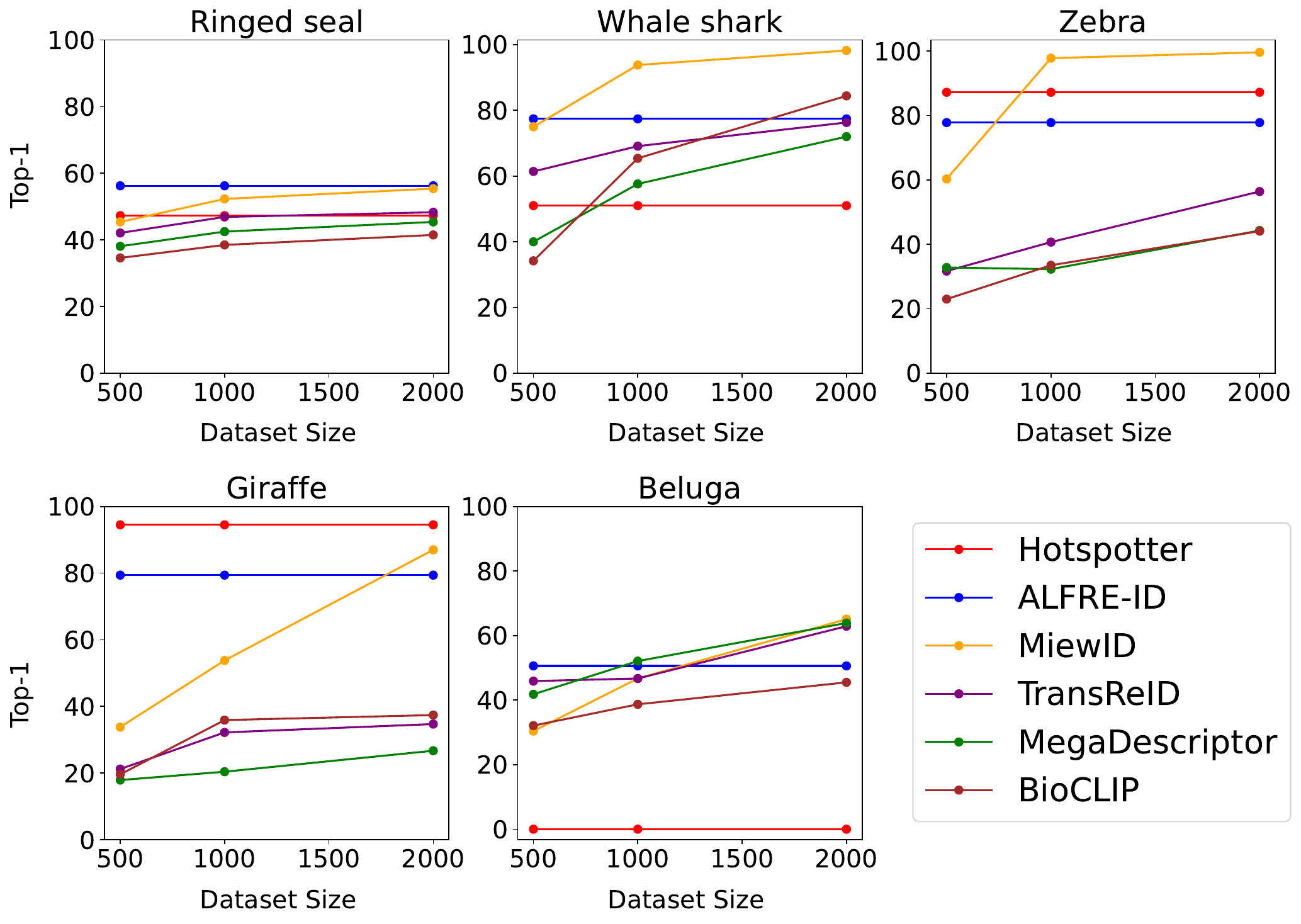}
    \caption{Dependency of Top-1 accuracies of methods on dataset size for each of five species.}
    \label{fig:results}
\end{figure}

The results are shown in Table~\ref{tab:results} and 
Fig.~\ref{fig:results}. As expected, when the amount of training data is large, the end-to-end learning-based methods outperform the local feature-based approaches. Especially, the methods originally developed for animal re-identification (MiewID and MegaDescriptor) provide high accuracies, with MiewID being overall the best method (78.4\% average top-1 accuracy over all datasets). ALFRE-ID provides the highest accuracy on Saimaa ringed seals that have notable variation on what parts of the pattern are visible emphasizing the benefits of feature aggregation. It is also good to note that ALFRE-ID is based on NORPPA (NOvel Ringed seal re-identification by Pelage Pattern Aggregation) method~\cite{nepovinnykh2024norppa} developed specifically for ringed seals. The Top-3 and Top-10 accuracies, along with a detailed analysis of each method's performance and consistency, are provided in the Supplementary Materials.

CNN-based MiewID clearly outperforms transformer-based TransReID, MegaDescriptor, and BioCLIP. TransReID performs closely to MiewID on whale shark dataset, given sufficient amount of samples. There are at least two potential reasons for this: (1) even when the full training sets are used, the amount of images is still relatively low for transformer architectures with a notably higher number of parameters than CNNs, and (2) while transformers excel at modeling the global relations in an image (pattern), they are not as good at modeling local regions essential for the re-identification task. BioCLIP model utilizes lower ($224\times224$) image input size and was initially trained for species and higher level classification, which can be a reason of under-performing in re-identification tasks. 

Out of the local feature-based methods, ALFRE-ID outperforms Hotspotter in most of the datasets, suggesting that the CNN-based local features combined with the feature aggregation produce representative pattern embeddings. Hotspotter completely fails on Beluga whales since they lack a clear pattern. However, the other methods are able to extract enough information for reasonable re-identification accuracy.

More interesting results are observed with small training set sizes. Local feature-based methods are generally better when the training set contains only 500 images, ALFRE-ID providing the highest accuracy (61.9\% average top-1 accuracy over all datasets). One notable benefit of ALFRE-ID is that as it uses learnable CNN-based local features, it would be possible to fine-tune them if training data on point correspondences were available making it possible to also utilizes larger datasets efficiently. This, however, was not evaluated in this study, but only pre-trained local features were used. Any advantages of using transformer architectures, such as TransReID, MegaDescriptor and BioCLIP, as it was demonstrated, are not observed in the small dataset scenario.

The characteristics of the species appear to have a notable effect on how quickly the end-to-end learning-based methods surpass ALFRE-ID and Hotspotter when the quantity of training images is increased. Intra-individual variance appears to be particularly influential, as species such as zebras and whale sharks, which exhibit smaller pose variations and patterns with clear contrast, thus reducing the impact of varying illumination, require less training data.

As a possible direction for future work, we would like to add a quantitative measure to describe the complexity of different animal marking patterns and study the correlation between the complexity of these patterns and the performance of re-identification algorithms. Additionally, we plan to include other animal species and re-identification methods in our research.


\section{Conclusion}
\label{sec:conc}
In this study, we examined automated animal re-identification, focusing on the amount of training data required. Given the challenges in gathering and annotating image data of wildlife animals, particularly endangered species, the adequacy of data for training methods remains a significant hurdle. Therefore, understanding the variations in data requirements among re-identification approaches is essential for selecting the most effective method based on available resources. Our experiments indicate that local feature-based methods generally outperform other methods when labeled data is limited. Therefore, methods like ALFRE-ID, leveraging modern local features, emerge as an appealing choice for various species. Conversely, end-to-end learning-based methods surpass local feature-based ones when enough training data is available. However, the requirements for training data vary depending on species-specific characteristics, notably intra-individual variance. CNN-based MiewID outperformed transformer-based methods, underscoring the challenge of obtaining enough data for training single-species transformer-based models.

\section*{Acknowledgements}
The authors would like to thank Finnish Cultural Foundation, Raija ja Ossi Tuuliaisen Säätiö Foundation, the project CoExist (Project ID: KS1549) for funding the research. In addition, authors would like to thank Department of Environmental and Biological Sciences at the University of Eastern Finland (UEF), the Zoological Society of San Diego d/b/a San Diego Zoo Wildlife Alliance and Wild Me, a lab of Conservation X Labs, for providing the data and identifying each individual.

{
    \small
    \bibliographystyle{ieeenat_fullname}
    \bibliography{main}
}

\clearpage
\setcounter{page}{1}
\renewcommand{\thesection}{\Alph{section}.\arabic{section}}
\setcounter{section}{0}
\renewcommand{\thetable}{\Alph{table}.\arabic{table}}
\setcounter{table}{0}
\maketitlesupplementary

In this supplementary material, we study the consistency of various re-identification models using mean average precision and Top-1, Top-3, Top-5, and Top-10 accuracies. 
\section{Consistency analysis}
\label{sec:consistency}

Table~\ref{tab:results-extended} presents an extended version of Table~\ref{tab:results}, including Top-3 and Top-10 accuracies for each method.

Table~\ref{tab:results-extended} displays re-identification results for six different models: HotSpotter~\cite{hotspotter}, ALFRE-ID~\cite{nepovinnykh2024alfreid}, MiewID~\cite{miewid}, TransReID~\cite{transreid}, MegaDescriptor~\cite{vcermak2024wildlifedatasets}, and BioCLIP~\cite{stevens2024bioclip}, validated on five distinct animal species: Saimaa Ringed Seals, Whale Sharks, Grevy's zebras, Masai Giraffes, and Beluga Whales. Each model exhibits slightly different performance across different species and different training sizes.

The consistency of HotSpotter varies significantly across different animal species. While it shows high performance and consistency for Grevy's Zebra and Masai Giraffe, with accuracy increasing significantly for Top-3, Top-5, and Top-10, its performance is moderate for Saimaa Ringed Seal and Whale Shark, and it fails completely on Beluga Whale.

ALFRE-ID shows relatively high performance for Saimaa Ringed Seal, Whale Shark, Grevy's Zebra, and Masai Giraffe, with Top-1 to Top-10 accuracies ranging from the mid-60s to the low-80s and mAP scores between 26.1\% and 49.1\%. This indicates a consistent ability to accurately re-identify individuals within these species. The performance is moderate for Beluga Whale which can be explained by the fact that Beluga whales do not have a distinguishable pelage pattern that can be used for re-identification.

MiewID shows excellent performance with full training data, particularly for Whale Shark and Grevy's Zebra, achieving near-perfect accuracy and mAP scores. The model shows moderate performance for Saimaa Ringed Seal but it is still reasonably consistent. The accuracy of MiewID is high when enough training data is available, however, the model's performance degrades significantly as the amount of training data decreases, particularly for Masai Giraffe and Beluga Whale. This indicates that while MiewID is highly effective with sufficient data, its robustness and consistency are compromised under smaller dataset scenarios.

The consistency of TransReID varies significantly across different species and training data sizes. The model performs quite well for Whale Sharks, Grevy's Zebra and Beluga Whale with some resilience to reduced data but noticeable performance dips. Saimaa Ringed Seal and Masai Giraffe exhibit less consistency and robustness, with significant performance drops as training data decreases. 

MegaDescriptor has a good consistency and low accuracy drop off on Saimaa Ringed Seal, but below average performance. The rest of the datasets under consideration show consistent decrease of accuracy over lower number of samples, with Whale sharks being the most impacting. Overall, the model demonstrates greater stability in comparison to other vision transformers, but falls short in achieving high accuracy.

BioCLIP's consistency also shows a variation depending on the species and the amount of training data used. The model performs well only for Whale Sharks, but the accuracy is quite low for the rest of the data with significant performance drops as training data decreases. These results can be explained by the fact that the input size of the BioCLIP model is only 224x224 pixels, and the Whale Shark dataset is the only dataset cropped to squared bounding boxes containing a fin and the surrounding area. If the BioCLIP model were improved to accept larger input data, it would likely show much better performance on full-size datasets.

\newgeometry{margin=1cm}
\thispagestyle{empty}

\begin{sidewaystable*}[h!]
\centering
\caption{Re-identification results. The highest top-1 accuracies for 0/500, 1000, 2000, and full size training sets are bolded.}
\begin{footnotesize}
\begin{tblr}{
  cell{1}{1} = {r=2}{valign=m},
  cell{1}{2} = {r=2}{valign=m},
  cell{1}{3} = {c=5}{c},
  cell{1}{8} = {c=5}{c},
  cell{1}{13} = {c=5}{c},
  cell{1}{18} = {c=5}{c},
  cell{1}{23} = {c=5}{c},
  cell{5}{1} = {r=4}{},
  cell{9}{1} = {r=4}{},
  cell{13}{1} = {r=4}{},
  cell{17}{1} = {r=4}{},
  vlines,
  hline{1,3-5,9,13,17,21} = {-}{},
  hline{2} = {3-27}{},
  hline{6-8,10-12,14-16,18-20} = {2-27}{},
  hspan=minimal,
  colsep=4pt
}
\begin{sideways}Method\end{sideways}                                   
& \begin{sideways}Training\hspace{6pt} data \vspace{-3pt}\end{sideways} 
& Saimaa ringed seal & & & & & Whale shark & & & & & Grevy's zebra & & & & & Masai giraffe & & & & & Beluga whale  & & & & \\
& & Top-1 & Top-3 & Top-5 & Top-10 & mAP & Top-1 & Top-3 & Top-5 & Top-10 & mAP & Top-1 & Top-3 & Top-5 & Top-10 & mAP & Top-1 & Top-3 & Top-5 & Top-10 & mAP & Top-1 & Top-3 & Top-5 & Top-10 & mAP \\
\begin{sideways}Hotspotter~\cite{hotspotter}\vspace{-3pt}\end{sideways} 
& 0 & 44.3 & 46.0 & 47.3 & 51.1 & 18.6 & 49.45 & 50.1 & 51.0 & 51.9 & 27.4 & \textbf{78.4} & 83.5 & 87.2 & 89.7 & 82.6 & \textbf{84.9} & 90.8 & 94.5 & 95.7 & 38.2 & 0.0 & 0.0 & 0.0 & 0.0 & 0.0 \\
\begin{sideways}ALFRE-ID~\cite{nepovinnykh2024alfreid}\vspace{-3pt}\end{sideways}   
& 0 & \textbf{53.9} & 55.8 & 56.3 & 64.5 & 33.2 & \textbf{73.5} & 76.7 & 77.4 & 78.0 & 49.1 & 69.6 & 75.0 & 77.8 & 81.6 & 40.5 & 72.3 & 77.6 & 79.4 & 80.7 & 26.1 & \textbf{40.4} & 48.8 & 50.6 & 55.5 & 22.2 \\
\begin{sideways}MiewID~\cite{miewid}\end{sideways}                                   
& full & \textbf{43.8} & 52.3 & 55.4 & 60.8 & 30.8 & \textbf{95.0} & 97.7 & 97.9 & 98.5 & 90.3 & \textbf{99.1} & 99.3 & 99.4 & 99.4 & 99.2 & \textbf{84.3} & 86.5 & 91.0 & 92.7 & 87.5 & \textbf{69.8} & 75.4 & 79.4 & 82.3 & 67.5 \\
& 2000 & \textbf{43.8} & 52.3 & 55.4 & 60.8 & 30.8 & \textbf{96.5} & 97.3 & 98.2 & 98.5 & 92.0 & \textbf{98.7} & 98.8 & 99.6 & 99.6 & 99.1 & \textbf{76.4} & 84.4 & 87.0 & 90.1 & 81.2 & \textbf{46.7} & 60.5 & 65.1 & 71.1 & 41.8 \\
& 1000 & \textbf{40.0} & 47.3 & 52.3 & 56.2 & 27.8 & \textbf{86.9} & 92.0 & 93.8 & 95.0 & 72.2 & \textbf{95.8} & 97.5 & 97.8 & 98.3 & 96.7 & \textbf{28.4} & 46.3 & 53.8 & 62.7 & 40.5 & 30.8 & 40.9 & 46.7 & 55.7 & 25.5 \\
& 500 & 33.1 & 41.5 & 45.4 & 52.5 & 23.1 & 58.4 & 70.3 & 75.0 & 80.8 & 31.8 & 46.4 & 54.5 & 60.3 & 66.8 & 53.1 & 2.1 & 27.8 & 33.8 & 42.7 & 17.7 & 16.2 & 25.6 & 30.4 & 39.2 & 13.4 \\
\begin{sideways}TransReID~\cite{transreid}\end{sideways}
& full & 37.7 & 44.4 & 48.3 & 56.5 & 25.5 & 93.2 & 95.8 & 96.7 & 97.6 & 84.3 & 45.2 & 54.9 & 60.4 & 67.8 & 52.8 & 18.0 & 31.8 & 38.7 & 48.6 & 28.2 & 38.5 & 54.3 & 59.8 & 67.7 & 34.4\\
& 2000 & 37.7 & 44.4 & 48.3 & 56.5 & 25.5 & 52.2 & 70.1 & 76.3 & 83.5 & 32.3 & 42.2 & 51.6 & 56.4 & 63.6 & 49.4 & 15.0 & 28.2 & 34.7 & 44.8 & 25.1 & 42.6 & 57.6 & 62.9 & 70.4 & 36.7\\
& 1000 & 34.8 & 42.7 & 46.9 & 54.0 & 25.3 & 43.9 & 61.4 & 69.1 & 77.8 & 23.9 & 27.1 & 35.8 & 40.7 & 47.1 & 34.0 & 17.3 & 26.7 & 32.2 & 40.4 & 25.1 & 25.9 & 38.1 & 46.7 & 59.5 & 23.7\\
& 500 & 30.6 & 40.0 & 42.1 & 48.5 & 21.1 & 37.8 & 53.5 & 61.4 & 71.0 & 19.3 & 22.3 & 28.7 & 31.7 & 37.2 & 27.6 & 9.7 & 16.6 & 21.2 & 26.8 & 15.4 & 22.5 & 37.6 & 45.9 & 55.3 & 21.5\\
\begin{sideways}MegaDescriptor~\cite{vcermak2024wildlifedatasets}\end{sideways}
& full & 32.9 & 40.4 & 43.3 & 48.1 & 22.9 & 86.7 & 91.7 & 93.7 & 96.7 & 74.7 & 40.6 & 49.3 & 53.9 & 61.3 & 47.3 & 15.7 & 23.0 & 27.9 & 36.2 & 22.4 & 44.2 & 56.5 & 61.5 & 71.5 & 39.7 \\
& 2000 & 32.9 & 40.4 & 43.3 & 48.1 & 22.9 & 45.8 & 62.4 & 72.0 & 80.2 & 28.0 & 32.0 & 40.6 & 44.3 & 49.9 & 38.5 & 14.5 & 22.2 & 26.7 & 34.2 & 21.3 & 43.0 & 57.9 & 63.9 & 73.4 & 39.4 \\
& 1000 & 30.4 & 37.7 & 42.5 & 47.9 & 21.9 & 32.5 & 48.4 & 57.6 & 67.8 & 16.7 & 23.6 & 30.4 & 32.3 & 37.0 & 28.8 & 11.0 & 16.7 & 20.4 & 26.8 & 16.4 & \textbf{33.5} & 45.4 & 52.1 & 60.5 & 29.0 \\
& 500 & 28.1 & 36.0 & 38.1 & 44.6 & 19.3 & 19.4 & 33.1 & 40.0 & 51.8 & 8.6 & 24.6 & 28.7 & 32.8 & 37.1 & 28.8 & 9.9 & 14.7 & 17.9 & 23.2 & 14.7 & 24.6 & 35.2 & 41.8 & 52.7 & 20.9 \\
\begin{sideways}BioCLIP~\cite{stevens2024bioclip}\end{sideways}
& full & 29.4 & 38.1 & 41.5 & 45.6 & 19.5 & 86.3 & 92.7 & 95.2 & 96.7 & 75.3 & 32.5 & 43.3 & 48.6 & 57.5 & 40.9 & 18.9 & 30.1 & 36.0 & 46.5 & 28.2 & 23.0 & 36.6 & 43.3 & 55.0 & 22.8\\
& 2000 & 29.4 & 38.1 & 41.5 & 45.6 & 19.5 & 65.5 & 79.0 & 84.4 & 88.9 & 45.2 & 29.7 & 37.8 & 44.1 & 52.5 & 36.9 & 19.2 & 30.2 & 37.4 & 49.0 & 28.6 & 25.6 & 38.7 & 45.5 & 55.2 & 22.1\\
& 1000 & 27.1 & 34.6 & 38.5 & 44.2 & 17.9 & 40.9 & 58.3 & 65.4 & 76.2 & 24.1 & 23.0 & 30.1 & 33.5 & 38.4 & 28.7 & 18.6 & 30.2 & 35.9 & 45.7 & 27.9 & 18.0 & 31.1 & 38.7 & 48.3 & 16.2\\
& 500 & 25.4 & 31.9 & 34.6 & 39.0 & 16.4 & 13.2 & 26.4 & 34.2 & 46.3 & 7.0 & 15.2 & 20.0 & 23.0 & 26.5 & 19.5 & 9.1 & 16.0 & 19.6 & 27.3 & 15.2 & 15.6 & 26.5 & 32.1 & 42.3 & 14.5
\end{tblr}
\end{footnotesize}
\label{tab:results-extended}
\end{sidewaystable*}
\restoregeometry

\clearpage
\thispagestyle{empty}




\end{document}